\documentclass[review]{elsarticle}
\journal{Neurocomputing}

\usepackage[utf8]{inputenc} 
\usepackage[T1]{fontenc}    
\usepackage{hyperref}       
\usepackage{url}            
\usepackage{booktabs}       
\usepackage{amsfonts}       
\usepackage{nicefrac}       
\usepackage{microtype}      
\usepackage{lipsum}
\usepackage{graphicx}
\usepackage{multirow}
\usepackage{stackengine}
\usepackage[table]{xcolor}
\usepackage{array}
\usepackage{framed,multirow}
\usepackage{amssymb} 
\usepackage{amsmath,amssymb}
\usepackage{color}
\usepackage{subfig}
\usepackage{caption}
\usepackage[ruled, lined, linesnumbered, commentsnumbered, longend]{algorithm2e}
\usepackage{algpseudocode, algcompatible}
\usepackage{mathtools}
\usepackage{natbib}

\begin{document}
\begin{frontmatter}

\title{Learning from Multiple Expert Annotators for Enhancing Anomaly Detection in Medical Image Analysis}

\author{Khiem H. Le$^{1,\dag}$, Tuan V. Tran$^{1,\dag}$, Hieu H. Pham$^{1,2,3,*}$\\Hieu T. Nguyen$^{1}$, Tung T. Le$^{1}$, Ha Q. Nguyen$^{1,2}$}
\address{$^{1}$Smart Health Center, VinBigData JSC, Hanoi, Vietnam \\ $^{2}$College of Engineering and Computer Science, VinUniversity, Hanoi, Vietnam \\
$^{3}$VinUni-Illinois Smart Health Center, Hanoi, Vietnam\\[-1cm]}

\cortext[mycorrespondingauthor]{Corresponding author:\textcolor{blue}{\texttt{ \underline{hieu.ph@vinuni.edu.vn}}}  (Hieu H. Pham)\\
\hspace*{0.3cm} $\dag$ These authors share the first authorship of this paper.}
\begin{abstract}
Building an accurate computer-aided diagnosis system based on data-driven approaches requires a large amount of high-quality labeled data. In medical imaging analysis, multiple expert annotators often produce subjective estimates about ``ground truth labels'' during the annotation process, depending on their expertise and experience. As a result, the labeled data may contain a variety of human biases with a high rate of disagreement among annotators, which significantly affect the performance of supervised machine learning algorithms. To tackle this challenge, we propose a simple yet effective approach to combine annotations from multiple radiology experts for training a deep learning-based detector that aims to detect abnormalities on medical scans. The proposed method first estimates the ground truth annotations and confidence scores of training examples. The estimated annotations and their scores are then used to train a deep learning detector with a re-weighted loss function to localize abnormal findings. We conduct an extensive experimental evaluation of the proposed approach on both simulated and real-world medical imaging datasets. The experimental results show that our approach significantly outperforms baseline approaches that do not consider the disagreements among annotators, including methods in which all of the noisy annotations are treated equally as ground truth and the ensemble of different models trained on different label sets provided separately by annotators.
\end{abstract}
\begin{keyword}
Supervised learning \sep
multiple annotators \sep object detection.
\end{keyword}
\end{frontmatter}

\section{Introduction}

Computer-aided diagnosis (CAD) systems for medical imaging analysis are getting more and more successful thanks to the availability of large-scale labeled datasets and the advances of supervised learning algorithms \cite{litjens2017survey,chan2020computer}. To reach expert-level performance, those algorithms usually require high-quality label sets, commonly scarce because of the costly and intensive labeling procedures. A typical label collection process in medical imaging is ``\textit{repeated-labeling}'', where multiple clinical experts annotate each data instance to overcome human biases \cite{razzak2018deep,irvin2019chexpert,nguyen2020vindr}. However, because of the differences from annotator biases and proficiency, annotations from the repeated-labeling process often suffer from high inter-reader variability \cite{pmid23220902,pmid24059400,pmid16569780}, which could reduce leaning performance if we treat them as ground-truth.

Many prior works have been done to mitigate inter-reader variations in annotations, which can be categorized into two main groups: (i) one-stage approach and (ii) two-stage approach. The first group learns the model, annotators' proficiency, and latent true labels jointly. Meanwhile, the second group first estimates the true label of each instance from its multiple label sets~\cite{Sheng_Zhang_2019}. This process is known as ``\textit{truth inference}''. After that, a supervised learning model is trained on the estimated true labels. All of those approaches show impressive results on both classification and segmentation problems \cite{tanno2019learning,NEURIPS2020_b5d17ed2}.

This work aims at addressing a fundamental question ``\textit{How to train a deep learning-based detector effectively from a set of possibly noisy labeled data provided by multiple annotators?}''~\cite{rodrigues2018deep}. To this end, we introduce a novel approach that learns from multiple expert annotators to improve the performance of a deep neural network in detecting abnormalities from chest X-ray images. The proposed approach, as visualized in Figure \ref{fig:main_diagram}, consists of two stages. The first one is truth inference using Weighted Boxes Fusion (WBF) algorithm~\cite{Solovyev_2021} to estimate the true labels and their confidence scores. The second stage is to train an object detector on estimated labels with a re-weighted loss function using implicit annotators' agreement, which is represented by the estimated confidence scores. For evaluation, we first simulate and test the proposed approach on a multiple-experts-detection dataset from MNIST~\cite{lecun2010mnist} called MED-MNIST. We then validate our approach on a real-world chest X-ray dataset with radiologist's annotations. Experiments on those scenarios demonstrate that the proposed approach provides better detection performance in terms of mAP scores than the baseline of treating multiple annotations as ground truth and the ensemble of models supervised by individual expert annotations.\\
\\
In summary, our main contributions in this work are two-folds: 
\begin{itemize}
  \item First, we introduce a simple yet effective method that allows a deep learning network to learn from multiple annotators to improve its performance in detecting abnormalities from medical images. The proposed approach aims at estimating the true annotations from multiple experts with confidence scores and uses these annotations to train a deep learning-based detector. This helps remove uncertainty in the learning process and provides higher label quality to train predictive models. 
  
  \item Second, the proposed approach demonstrates its effectiveness on both simulated and real medical imaging datasets by surpassing current state-of-the-art methods on the context of learning with multiple annotators. In particular, our method is simple and can be applied for a wide range of applications in medical imaging and object detection in general. The codes used in the experiments are available on our Github page at \url{https://github.com/huyhieupham/learning-from-multiple-annotators}. We also have made the dataset used in this study available for public access on our project’s webpage at \url{https://vindr.ai/datasets/cxr}. 
\end{itemize}

The rest of the paper is organized as follows. Related works on learning from multiple annotators and weighted training techniques are reviewed in Section~\ref{sec:2}. Section~\ref{sec:3} presents the details of the proposed method with a focus on how to estimate the ground truth annotations from multiple experts. Section~\ref{sec:4} provides comprehensive experiments on a simulated object detection dataset and a real-world chest X-ray dataset. Section~\ref{sec:5} discusses the experimental results, some key findings, and limitations of this work. Finally, Section~\ref{sec:6} concludes the paper.

\section{Related works}
\label{sec:2}

\textbf{Learning from multiple annotators}. There are two major lines of research on learning from multiple annotators: two-stage approaches~\cite{10.14778/3055540.3055547,Sheng_Zhang_2019, jin2020technical} and one-stage approaches~\cite{tanno2019learning,li2021learning,HumanError2020}. Two-stage approaches infer the true labels first, then train a model using the estimated ones. The most simple solution for label aggregation is majority voting, in which the choice of majority annotators regards as the truth~\cite{10.1145/1401890.1401965}. However, when the skill levels of the annotators differ, the majority voting strategy may not work well. This is a common occurrence in the general ``\textit{learning from crowds}'' problem when ``spammers'' are present. Later approaches typically incorporate other information into the truth inference procedure, such as the annotators' proficiency~\cite{NIPS2011_c667d53a}, annotators' confusion matrix~\cite{Dawid1979MaximumLE,10.5555/2998687.2998822}, or the difficulty of each sample \cite{whitehill2009whose}. While two-stage approaches have the advantage of simplicity in both implementing and debugging, they do not make use of the raw annotations in model learning. One-stage approaches address this issue by simultaneously estimating the hidden true labels and learning the desired model from noisy labels of multiple annotators. Earlier works in this group use Expectation Maximization (EM) algorithm~\cite{10.1145/1553374.1553488} for jointly modeling the annotators' ability and the latent ground-truth. More recent approaches employ end-to-end frameworks which enable the neural networks to learn directly from the noisy labels~\cite{rodrigues2018deep}, and further developed by incorporating annotators' confusion matrix~\cite{NEURIPS2020_b5d17ed2,tanno2019learning}, or instance features~\cite{li2021learning}.

\textbf{Weighted training examples}. In this paper, we propose a new re-weighted loss function in which we assign more weights to examples that we consider be more confident. Previous works on the use of weighted training examples can be briefly categorized into two groups: (i) emphasize hard examples and (ii) emphasize easy examples. Methods in the group (i) include hard-example mining~\cite{1467360,shrivastava2016training}, which is a bootstrapping technique over the difficult examples; boosting algorithms~\cite{schapire2013explaining}, where the misclassified examples in preceding weak classifiers are assigned with higher weights; and focal  loss~\cite{lin2017focal} that addresses class imbalance problems by adding a regulator to the cross-entropy loss for focusing on hard negative examples. Works in the group (ii) are instances of broader topics such as curriculum learning~\cite{10.1145/1553374.1553380}, which is biologically inspired by human gradual learning, with easier examples are preferred in early training stages; learning with noisy labels~\cite{zhang2021understanding,10.5555/3305381.3305406}, which prefers examples with smaller training losses as they are more likely to be clean.

Unlike any approaches above, we propose in this paper a new loss function that assigns more weights to more confident examples that determine by the consensus of multiple annotators. Our experimental results validate the correctness of this hypothesis.   

\section{Proposed Method}

This section presents details of the proposed method. We first give a formulation on learning from multiple annotators (Section~\ref{setc.3.1}). We then introduce a simple way to estimate the true labels from multiple annotators (Section~\ref{setc.3.2}). Next, our network architecture and training methodology with a new re-weighted loss function are described (Section~\ref{setc.3.3}).

\label{sec:3}
\label{sec:method}
\subsection{Problem formulation}
\label{setc.3.1}
Given a set of $N$ training images $\left\{\mathbf{x}_{i}\right\}_{i=1}^{N}$ with corresponding bounding box annotations $\left\{\tilde{y}_{i}^{(r)}\right\}_{i=1}^{N}$ from multiple annotators where $\tilde{y}_{i}^{(r)}$ denotes the label for example $\mathbf{x}_{i}$ given by annotator $r \in \mathbb{ S}(R)$, which ${\mathbb S}(R)$ is a set of $R$ different expert annotators. In this study, we make use of those expert annotations $\left\{\tilde{y}_{i}^{(r)}\right\}_{i=1}^{N}$ to estimate a single set of true labels with confidence scores $\left\{\mathbf{y}_{i}; c_i\right\}_{i=1}^{N}$. We then train a supervised object detector with the estimated labels using the proposed re-weighted loss function. In order to evaluate the effectiveness of the proposed method, we use a gold-standard test set $\mathcal{T}=\left\{\left(\mathbf{x}^{(j)}, \mathbf{y}^{(j)}\right)\right\}_{j=1}^{M}$ containing $M$ examples. In medical imaging scenarios, where the true labels are not available, we obtain the gold-standard test labels $\mathbf{y}^{(j)}$ from the consensus of a group of experiences radiologists. Figure~\ref{fig:main_diagram} below shows an overview of the proposed method.

\begin{figure}[ht]
  \centering
  \includegraphics[width=1.0\textwidth]{{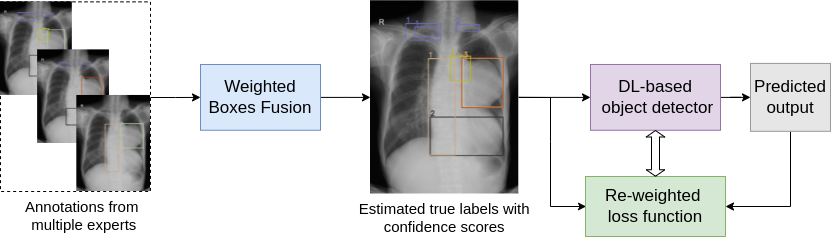}}
  \caption{Illustration of the proposed approach that aims to build a deep learning system for abnormal detection on medical scans from multiple expert annotators. The training process contains two stages. The first stage focuses on truth inference, in which it estimates the true labels using the WBF algorithm~\cite{Solovyev_2021} with the implicit annotator’s agreement as confidence scores. The second uses the estimated confidence scores to train a deep learning-based detector using a re-weighted object detection loss function. To provide abnormality analysis during the testing phase, only the fully trained image detector is required.}
  \label{fig:main_diagram}
\end{figure} 

\subsection{Estimating the true labels from multiple expert annotators}
\label{setc.3.2}

\begin{figure}[ht]
  \centering
  \subfloat[The original annotations provided by multiple radiology experts. The same abnormal finding is represented by the sample color.]{%
  \includegraphics[width=0.45\textwidth]{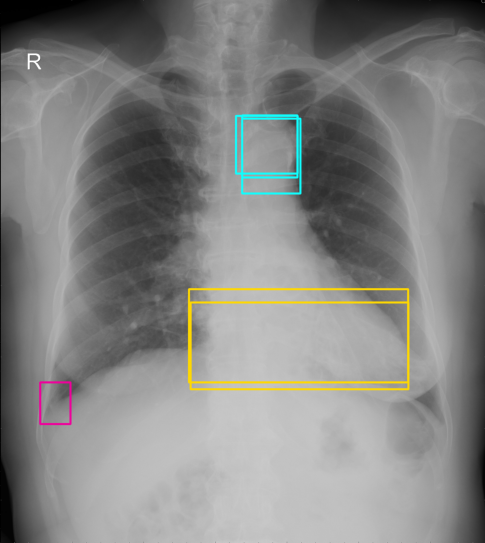}%
  }\hfill
  \subfloat[Fused boxes with corresponding confidence scores after applied the WBF algorithm.]{%
  \includegraphics[width=0.45\textwidth]{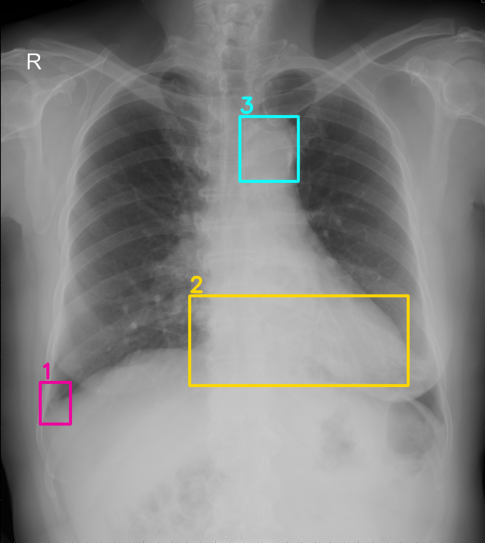}%
  }\hfill
  \caption{(a) Visualization of multiple expert annotations on a chest X-ray example from the VinDr-CXR dataset~\cite{nguyen2020vindr} and (b) the fused boxes with confidence scores obtained by the WBF algorithm.}
  \label{fig:wbf}
\end{figure}

We firstly estimate the true labels using Weighted Boxes Fusion (WBF) algorithm~\cite{Solovyev_2021}. This technique is used for combining predictions from multiple sources, i.e., using ensemble to achieve better prediction results or combining labels of different expert annotators. We describe the WBF algorithm in more detail in Algorithm \ref{alg:wbf}. The final examples used to train deep learning detectors contain merged boxes with confidence scores. The visualization of fused boxes and the corresponding confidence scores are shown in Figure~\ref{fig:wbf}. Our fusion box algorithm emphasizes that the greater agreement between bounding boxes (e.g., two or three annotators have the same diagnosis for an abnormal finding on the image), the more likely the box annotation is correct.

\subsection{Network architecture and training methodology} 
\label{setc.3.3}

\begin{algorithm}
\begin{footnotesize}
\caption{\textbf{The WBF algorithm applied for multiple expert annotations}}
\label{alg:wbf}

\KwIn{An image $\mathbf{x}$ with a list of annotations $\tilde{y}$ given by a set $\mathbb{S}(R)$ of $R$ experts. The expert $r \in \mathbb{S}(R)$ with proficiency $p_{r}$ provides the annotations including $r_x$ boxes, $A_r = \left[  \text{box}_1, \ldots, \text{box}_{r_x}\right]$. All of the experts' annotations being merged into a list $A$.

}

\KwOut{A list of $k$ fused boxes $F = \left[  \text{box}_1, \ldots, \text{box}_{k}\right]$.}

\BlankLine
Declare empty lists $L$ and $F$ for boxes clusters and fused boxes, respectively. Each position in the list $L$ can have a cluster of boxes or a single box. Each position in $F$ has only one box, which is the fused box from the corresponding cluster in $L$.
\BlankLine \label{loop}
Iterate through all boxes in $A$ in a cycle and attempt to find a matching box in the list $F$. Two boxes are defined matched if they have a high degree of overlap $(\text{e.g. IoU} > 0.4)$. If there are more than one matching boxes in $F$, the one with the highest IoU will be chosen.
\BlankLine
If the matching box is not found in step \ref{loop}, add the current box to $L$ and $F$ as new entry for the new cluster before moving on to the next box in the list $A$.
\BlankLine
If the match is found in step \ref{loop}, add this box to the list $L$ at the position $\textit{pos}$ which corresponds to the matching box in the list $F$
\BlankLine
Set the fused box's coordinates $F[\textit{pos}]$ to be the weighted average of $T$ boxes accumulated in cluster $L[\textit{pos}]$ with the following formulas:
$$x_{1,2} \coloneqq \frac{\sum_{i=1}^{T} p_{i} x_{1,2}}{\sum_{i=1}^{T} p_{i}}$$
$$y_{1,2}  \coloneqq \frac{\sum_{i=1}^{T} p_{i} y_{1,2}}{\sum_{i=1}^{T} p_{i}}$$
\BlankLine
Set the the fused boxes' confidence scores in $F$ to the number of boxes in the corresponding cluster in $L$ once all boxes in $A$ have been processed. 
$$
c  \coloneqq c \min{(T, N)}
$$
The fused boxes with confidence scores now represent the annotators' level of agreement.
\end{footnotesize}
\end{algorithm}

Object detection is a multi-task problem, in which the loss function consists of two parts: (1) the localization loss $\mathcal{L}_\text{loc}$ for predicting bounding box offsets and (2) the classification loss $\mathcal{L}_\text{cls}$ for predicting conditional class probabilities. In this work, we focus on one-stage anchor-based detectors. A general form of the loss function for those detectors can be written as 

\begin{gather}
\label{eq:general_loss}
\begin{aligned}
\mathcal{L}\left(p, p^{*}, t, t^{*}\right) &=\mathcal{L}_{cls}\left(p, p^{*}\right)+\beta I(t) \mathcal{L}_{loc}\left(t, t^{*}\right) \\
I(t) &=\left\{\begin{array}{ll}
1 & \text {if} \operatorname{IoU}\left\{a, a^{*}\right\}>\eta \\
0 & \text { otherwise},
\end{array}\right.
\end{aligned}
\end{gather}
where $t$ and $t^*$ are the predicted and ground truth box coordinates, $p$ and $p^*$ are the class category probabilities, respectively; $\operatorname{IoU}\left\{a, a^{*}\right\}$ denotes the Intersection over Union (IoU) between the anchor $a$ and its ground truth $a^{*}$; $\eta$ is an IoU threshold for objectness, i.e. the confidence score of whether there is an object or not; $\beta$ is a constant for balancing two loss terms $\mathcal{L}_{cls}$ and $\mathcal{L}_{loc}$ \cite{zou2019object}.

We use fused boxes confidence scores $c^i_k$ obtained from Algorithm \ref{alg:wbf} to get a re-weighted loss function that emphasizes boxes with high annotators agreement. The new loss function, which we name it as Experts Agreement Re-weighted Loss (EARL) can now be written as
\begin{equation}
\label{eq:earl}
\mathcal{L}\left(p, p^{*}, t, t^{*}\right) = c \mathcal{L}_{cls}\left(p, p^{*}\right) + c \beta I(t) \mathcal{L}_{loc}\left(t, t^{*}\right),
\end{equation}
where $c$ is the fused box confidence score.

\section{Experiments}
\label{sec:4}
We validate the proposed method in both synthetic and real-world scenarios: (1) the MED-MNIST, an object detection dataset, which was simulated from MNIST~\cite{lecun2010mnist} with multiple expert annotations; (2) VinDr-CXR~\cite{nguyen2020vindr}, a chest X-ray dataset with labels provided by multiple radiologists. In the following sections, we describe those two datasets and our experiment setup, as well as the experimental results. 

\subsection{Datasets}
\label{setc.4.1}
\subsubsection{MED-MNIST Dataset}
\begin{figure}[ht]
  \centering
  \subfloat[Original transition matrix]{%
  \includegraphics[width=0.45\textwidth]{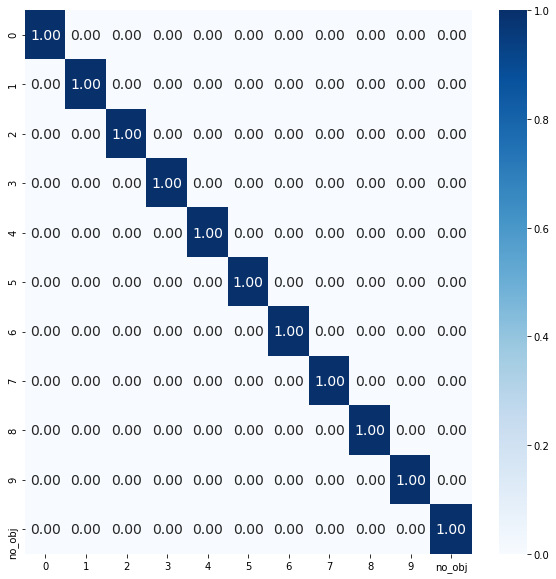}%
  }\hfill
  \subfloat[Simulated expert transition matrix]{%
  \includegraphics[width=0.45\textwidth]{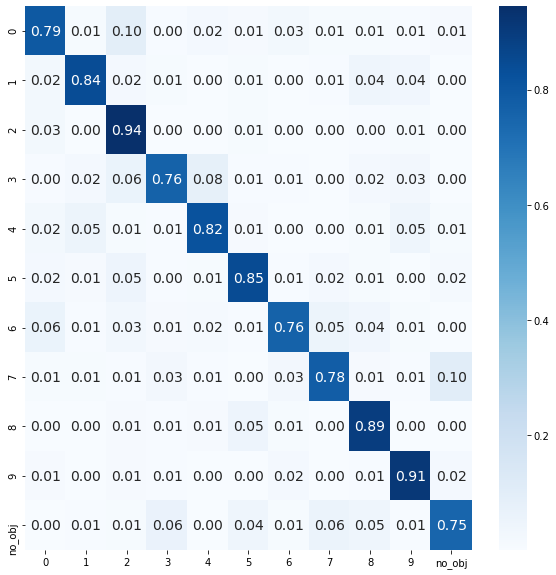}%
  }\hfill
  \caption{Visualization of the original and synthesized transition matrices. To simulate the false negative scenario, we use an additional class called \texttt{no\_obj}.}
  \label{fig:transition_matrices}
\end{figure}

\begin{figure}[ht]
  \centering
  \subfloat[MNIST Detection\label{det_mnist}]{%
  \includegraphics[width=0.45\textwidth]{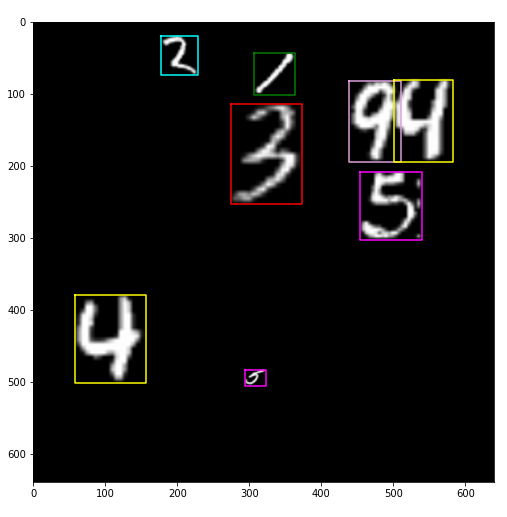}%
  }\hfill
  \subfloat[Simulated expert annotations\label{med_mnist}]{%
  \includegraphics[width=0.45\textwidth]{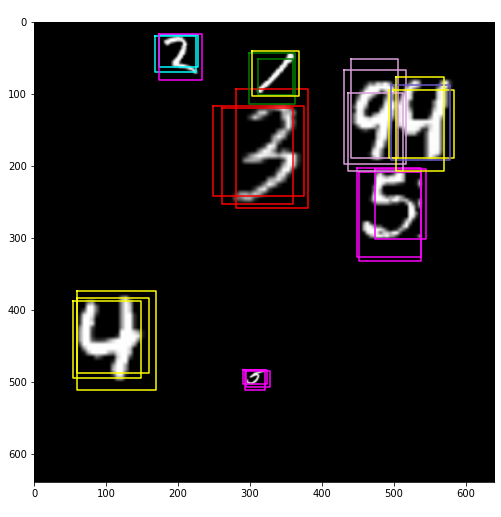}%
  }\hfill
  \caption{The MED-MNIST dataset with multiple expert annotations, obtained by perturbing boxes and classes from the MNIST dataset~\cite{lecun2010mnist}.}
  \label{fig:MED_MNIST}
\end{figure}

Based on MNIST~\cite{lecun2010mnist} -- a database of handwritten digits,  we synthesize a multiple-experts-detection dataset so-called MED-MNIST by two steps: (1) we construct the detection task by copying and pasting digits from MNIST into a black background with digit sizes are randomly chosen from a predefined range, the bounding box annotations would be the smallest rectangle that contains digits as visualized in Figure~\ref{det_mnist}; (2) we simulate $R$ different expert opinions for each sample, assuming those $R$ experts have the same proficiency $p$. The expert annotations are generated by varying two key factors that influence detection annotations: (i) class labels and (ii) object coordinates. To synthesize the expert annotations on class labels, we use an unique transition matrix $A_k  (k \in \{ 1,  \ldots, R\}) $ for each expert $E_k$ to compute probability distributions that represent the expert mis-classification. The proposed transition matrix is shown in Figure~\ref{fig:transition_matrices}. About the object coordinates, we simulate bounding box annotations that are highly overlapping with the true bounding box. Both factors (i) and (ii) are controlled by proficiency $p$. More specifically, $A_k$ are diagonally dominant $(a_{ii} > a_{ij} \text{ for all } i \ne j)$, and $a_{ii} = \text{min}(\text{max}(0.5, \alpha), 1) \text{ with } \alpha \sim \mathcal{N}(p, 0.05)\ $. The simulated bounding boxes are subject to have \textit{IoU} with the true bounding box being larger than $p$. Here we set the number of expert annotations per sample $R$ to 3, and the proficiency $p$ to $0.8$. The simulated MED-MNIST dataset consists of 5,000 samples for training, 1,000 for hold-out validation and 1,000 for testing.

\subsubsection{VinDr-CXR Dataset}
\begin{figure}[ht]
  \centering
  \subfloat{\includegraphics[width=\textwidth]{{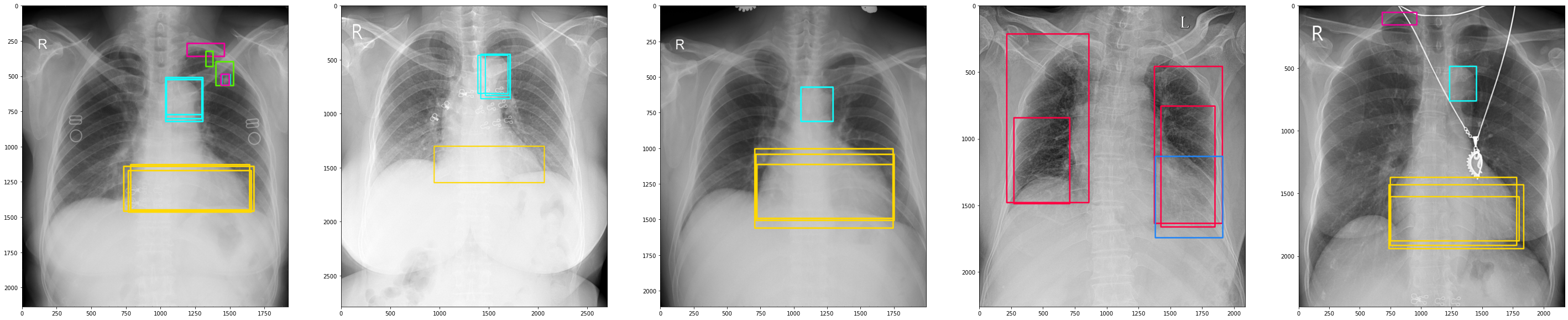}}}
  \\
  \subfloat{\includegraphics[width=\textwidth]{{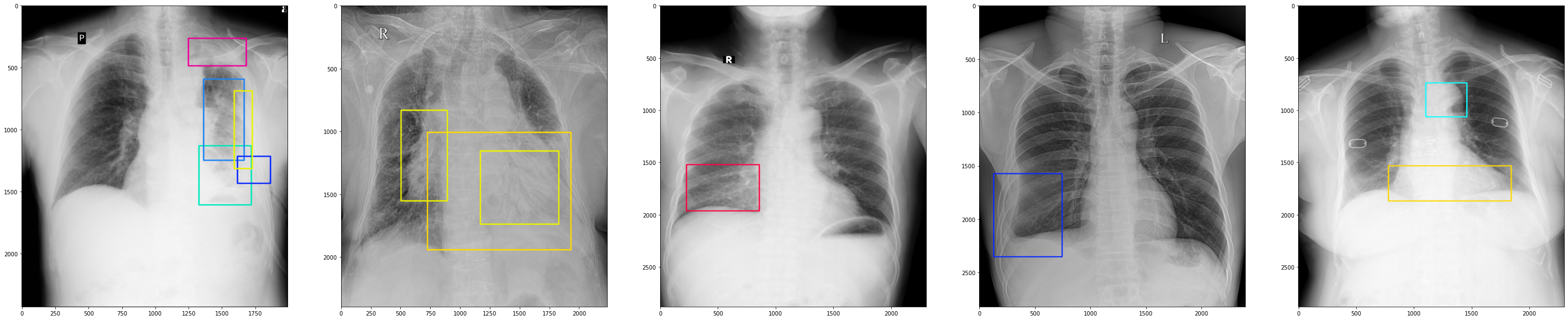}}}
  \caption{Visualization of abnormal findings (different bounding box colors represent different findings) from the VinDr-CXR dataset: (top) Each scan in the training set was annotated by three different radiologists; (bottom) Test set annotations were obtained from the consensus of five radiologists.}
  \label{fig:vindr_cxr}
\end{figure}

VinDr-CXR~\cite{nguyen2020vindr}, by far the largest public chest X-ray database with radiologist-generated annotations. It consists of 18,000 chest X-ray scans that come with both the localization of critical findings and the classification of common thoracic diseases. The dataset includes 15,000 scans for training and 3,000 scans for testing. In particular, the annotations were obtained by a group of 17 radiologists with at least eight years of experience. Each image in the training set was independently labeled by three radiologists, while the annotations in the test set were carefully treated and obtained by the consensus of 5 radiologists. Several examples from the VinDr-CXR dataset are shown in Figure~\ref{fig:vindr_cxr}. 

\subsubsection{Rads-VinDr-CXR Dataset}
One intriguing characteristic of the VinDr-CXR dataset~\cite{nguyen2020vindr} is that 94.28\% of the abnormal scans in the training set (3,315 out of 3,516) were annotated by a group of three radiologists with their correspondence IDs being \textit{R8}, \textit{R9} and \textit{R10}. As a result, we create Rads-VinDr-CXR, a sub-dataset that is only annotated by those three radiologists. The Rads-VinDr-CXR serves as a suitable multiple annotators dataset to validate the proposed approach.

\subsection{Experimental Setup}
\label{setc.4.2}

\subsubsection{Evaluation metric}

For all experiments, we report the detection performance using the standard mean average precision metric at a threshold of 0.4 (mAP@0.4)~\cite{everingham2010pascal}. Specifically, a predicted object is a true positive if it has an IoU of at least 0.4 with a ground truth bounding box. The average precision (AP) is the mean of 101 precision values, corresponding to recall values ranging from 0 to 1 with a step size of 0.01. The final metric is the mean of AP over all lesion categories. We also employ mAP@[0.5:0.95:0.05] as an additional metric to assess the model's performance on different IoU thresholds ranging from 0.5 to 0.95 with a step size of 0.05.

\subsubsection{Implementation Details}
The main detector used in our experiments is YOLOv5-S~\cite{glenn_jocher_2021_4679653}. The network is built with PyTorch 1.7.1 (\url{https://pytorch.org/}) and trained on two NVIDIA RTX 2080 Ti GPUs. All training and test images are resized to the dimension of $640 \times 640$ pixels. The detector is trained for 50 epochs with 1cycle learning rate decay~\cite{smith2017cyclical} using the SGD optimizer~\cite{ruder2016overview}. The initial learning rate is set to 1e-3.

To validate the robustness of the proposed approach across different deep learning detectors, we further train and evaluate EfficientDet~\cite{tan2020efficientdet} with sizes D3 and D4. Specifically, all images are resized to 640×640 pixels and the model is trained for 30 epochs with constant learning rate 3e-4 using the AdamW optimizer~\cite{loshchilov2017decoupled}.

\subsubsection{Comparison with the state-of-the-art}
To the best of our knowledge, there is no existing multiple-annotators model for object detection tasks in the literature. Hence, we compare the performance of the proposed method against the baseline, which uses all experts’ annotations per example without taking into account the disagreement among annotators. On the Rads-VinDr-CXR dataset, we further compare our method with the Rads-ensemble, which is the ensemble of independent models trained on separate radiologists' annotation sets. In this case, the WBF algorithm is used to combine the predictions of those models.

\subsection{Experimental Results}

Table \ref{tab:med_mnist} and Table \ref{tab:vindr_cxr_effdet} report the experimental results of the YOLOv5-S detector on MED-MNIST and VinDr-CXR datasets, respectively. On both synthetic and real-world datasets, the proposed approach outperforms the chosen baselines, even with the ensemble of individual experts' models. Specifically, on the test set of the MED-MNIST dataset, our method reports an overall mAP@0.4 of 0.980 and an overall mAP@[0.5:0.95:0.05] of 0.849. These results are much higher the performance of the baseline with mAP@0.4 = 0.975 and mAP@[0.5:0.95:0.05] = 0.815, boosting the mAP scores of the baseline by 0.51\% and 4.2\%, respectively. Experimental results on the VinDr-CXR and Rads-VinDr-CXR datasets also validate the effectiveness of the proposed method. We achieve an overall mAP@0.4 of 0.200 on the VinDr-CXR dataset and  an overall mAP@0.4 of 0.158 on the Rads-VinDr-CXR dataset. We emphasize that these results outperform both the baseline model, individual model trained on label provided by individual annotator (i.e. \textit{R8}, \textit{R9}, \textit{R10}), as well as the ensemble model. 

The experimental results with EfficientDet detector are provided in Table \ref{tab:vindr_cxr_effdet}. We found that better detection performances compared to the baseline have been reported. This evidence confirm the robustness of the proposed approach across deep learning detectors.

\begin{table}[ht]
\label{tab:med_mnist}
\centering
\caption{\label{tab:med_mnist} Experimental results on the MED-MNIST dataset. The highest scores are highlighted in \textcolor{red}{\textbf{red}}.}
\small{
\begin{tabular}{lcc}
\hline
\hline
\multicolumn{1}{c}{Method} & mAP@0.4 & mAP@{[}0.5:0.95:0.05{]} \\ \hline
Baseline                            & 0.975            & 0.815                         \\
WBF+EARL (ours)                            & \textcolor{red}{\textbf{0.980}}   & \textcolor{red}{\textbf{0.849}}                \\ \hline
\hline
\end{tabular}
}
\end{table}

\begin{table}[ht]
\centering
\caption{\label{tab:vindr_cxr} Experimental results on the VinDr-CXR and Rads-VinDr-CXR datasets with the YOLOv5-S detector. The highest scores are highlighted in \textcolor{red}{\textbf{red}}.}
\small{
 \begin{tabular}{lll}
\hline
\hline
Dataset                        & Method & mAP@0.4 \\ \hline
\multirow{2}{*}{VinDr-CXR}     & Baseline        & 0.190            \\
                               & WBF+EARL (ours)   & \textcolor{red}{\textbf{0.200}}   \\ \hline
\multirow{6}{*}{Rads-VinDr-CXR} & Baseline        & 0.148            \\
                               & R8              & 0.121            \\
                               & R9              & 0.132            \\
                               & R10             & 0.124            \\
                               & Rads-ensemble    & 0.154            \\
                               & WBF+EARL (ours)   & \textcolor{red}{\textbf{0.158}}   \\ \hline \hline
\end{tabular}
}
\end{table}

\begin{table}[ht]
\centering
\caption{\label{tab:vindr_cxr_effdet} Experimental results on the VinDr-CXR dataset while EfficientDet is used as the detector. The scores are measured in mAP@[0.5:0.95:0.05], with highest values highlighted in \textcolor{red}{\textbf{red}}.}
\small{
\begin{tabular}{lll}
\hline
\hline
                & Baseline & WBF+EARL \\ \hline
EfficientDet-D3 & 0.1142   & \textcolor{red}{\textbf{0.1353}}   \\
EfficientDet-D4 & 0.1223   & \textcolor{red}{\textbf{0.1431}}   \\ \hline \hline
\end{tabular}
}
\end{table}

\section{Discussions}
\label{sec:5}
\subsection{Key findings and meaning}
To the best of our knowledge, the proposed method is the first effort to train an image detector from labels provided by multiple annotators, which is crucial in constructing high-quality CAD systems for medical imaging analysis. In particular, we empirically showed a notable improvement in terms of mAP scores by estimating the true labels and then integrating the implicit annotators' agreement into the loss function to emphasize the clean bounding boxes over the noisy ones. The idea is simple but effective, allowing the overall framework can be applied in training any image machine learning-based detectors.

\subsection{Limitations}
Despite the higher predictive performance over the relevant baselines, we acknowledge that the proposed method has some limitations. First, the overall architecture is not end-to-end. It may not fully exploit the benefits of combining truth inference and training the desired image detector. Second, applying the WBF algorithm to annotation sets with a high level of noise may produce low-quality training data. This case is quite impractical in the medical imaging field when the annotators are experienced clinical experts, but it frequently occurs in the general \textit{learning from crowds} problems.

\section{Conclusion}
\label{sec:6}
This paper concentrates on the use of annotations from multiple experts to build a robust deep learning system for abnormality detection on medical images. We propose using Weighted Boxes Fusion (WBF) algorithm to obtain the aggregated annotations with the implicit annotators' agreement as confidence scores. The estimated annotations are then used to train a deep learning detector with a re-weighted loss function that incorporates the confidence scores to localize abnormal findings. We empirically demonstrate that the proposed approach outperforms current state-of-the-art baseline approaches in both synthetic and real-world scenarios. To the best of our knowledge, we introduce for the first time an effective method that trains an object detector from multiple annotators. We believe our method is simple and can be applied widely in medical imaging. 

\section{Acknowledgements} 
This work was supported by Smart Health Center at VinBigData JSC. The authors gratefully acknowledge all anonymous reviewers for their valuable comments and suggestions.
\bibliography{references}
\end{document}